\documentclass[10pt, a4paper, twocolumn]{IEEEtran}

\usepackage{amsmath,epsfig,amssymb,verbatim,amsopn,cite,multirow}
\usepackage{balance}
\usepackage{mathtools}
\usepackage[normalem]{ulem}
\usepackage{scalerel}
\usepackage[usenames,dvipsnames]{color}
\usepackage[all]{xy}  
\usepackage[bottom]{footmisc}
\usepackage{url}
\usepackage{amsfonts}
\usepackage{amssymb}
\usepackage{epsfig}
\usepackage{enumitem}
\usepackage{bm}
\usepackage{amsmath}
\usepackage{epsfig}
\usepackage{pdfsync}
\usepackage{amsmath,amssymb, amstext}
\usepackage{amsmath}
\usepackage{graphicx}
\usepackage{xcolor}
\usepackage{epstopdf}
\usepackage{algorithm}
\usepackage{algorithmicx }
\usepackage{algpseudocode}
\usepackage[justification=centering,font=scriptsize]{caption}
\usepackage{array}
\usepackage{amsmath}
\usepackage{amssymb}
\usepackage{amsthm}
\usepackage[utf8]{inputenc}
\usepackage{amsthm}
\usepackage[english]{babel}

\usepackage{xcolor}

\usepackage[left=.7in,right=.7in,top=1in,bottom=1in]{geometry}
\usepackage[export]{adjustbox}
\usepackage{graphicx}
\usepackage[utf8]{inputenc}
\usepackage[english]{babel}
\usepackage{xcolor}
\usepackage[T1]{fontenc}
\usepackage{subfig}
\usepackage[tableposition=top]{caption}
\usepackage{caption}
\usepackage{hhline}
\begin{document}
\bstctlcite{IEEEexample:BSTcontrol}

\title{An Anomaly Detection Method for Satellites Using Monte Carlo Dropout 
 } 

    \author{Mohammad Amin Maleki Sadr, Yeying Zhu,~\IEEEmembership{Member, IEEE,} 
    Peng Hu,~\IEEEmembership{Senior Member, IEEE}
\thanks{
This work was supported by the High-Throughput and
Secure Networks Challenge program of National Research
Council Canada under Grant No. CH-HTSN-418.
Mohammad Amin Maleki Sadr and Yeying Zhu are with the Dept. of Statistics and Actuarial Science, University of Waterloo, ON N2L 3G1, Canada. 
Peng Hu is with Digital Technologies Research Center, National Research Council Canada and with the Dept. of Statistics and Actuarial Science, University of Waterloo, ON N2L 3G1, Canada. (Email: Peng.Hu@nrc-cnrc.gc.ca)
}
}

\markboth{The Final Version Was Published in the IEEE Transactions on Aerospace and Electronic Systems on 22 SEP 2022. DOI:10.1109/TAES.2022.3206257}
{}

\maketitle
\begin{abstract}
Recently, there has been a significant amount of interest in satellite telemetry anomaly detection (AD) using neural networks (NN). For AD purposes, the current approaches focus on either forecasting or reconstruction of the time series, and they cannot measure the level of reliability or the probability of correct detection. Although the Bayesian neural network (BNN)-based approaches are well known for time series uncertainty estimation, they are computationally intractable. In this paper, we present a tractable approximation for BNN based on the Monte Carlo (MC) dropout method for capturing the uncertainty in the satellite telemetry time series, without sacrificing accuracy. For time series forecasting, we employ an NN, which consists of several Long Short-Term Memory (LSTM) layers followed by various dense layers. We employ the MC dropout inside each LSTM layer and before the dense layers for uncertainty estimation. With the proposed uncertainty region and by utilizing a post-processing filter, we can effectively capture the anomaly points. Numerical results show that our proposed time series AD approach outperforms the existing methods from both prediction accuracy and AD perspectives.
\end{abstract}
\begin{IEEEkeywords}
   Anomaly detection, telemetry time series  data, satellite communications, uncertainty estimation.
\end{IEEEkeywords}

\IEEEpeerreviewmaketitle
\section{Introduction}
Satellites are complicated systems composed of various interconnected technologies such as telemetry sensing, mobile communications, and navigation systems \cite{chen2021imbalanced}.  Proactive diagnosis of  failures, anomaly detection (AD), and response to potential hazards are required to guarantee the availability and continuity of satellite services \cite{li2019intelligent}. 
Considering the complex design structure and harsh  environment of the space, AD on the satellites cannot be performed
directly in the outer space; instead, the telemetry data is employed 
to provide  healthy key information that can be
used to recover  the satellite from  the possible problems \cite{safaeipour2021survey}. Previous satellite AD systems were based on expert knowledge. However, by receiving a large amount of data from different telemetry channels, the expert-based (human-based) AD  systems cannot properly discover anomalies, so data-driven intelligent AD  methods are recommended \cite{cayrac1996handling}. 

Many studies have focused on deep neural networks (DNN) for data-based AD  for spacecraft applications
\cite{schwabacher2009unsupervised,ibrahim2020machine}.   AD in this context has been studied from two different standpoints, which are: 
\subsubsection{Forecasting-based method}
In this approach,  the prediction error is computed as a
difference between the current and the predicted
 states. Then, by leveraging a thresholding-based method, an anomaly can be detected. The idea behind this approach is that an efficient detection method should
learn the expected behavior of a telemetry channel, so any deviation from the expected response can be flagged as a tentative anomaly \cite{hundman2018detecting, ding2018multivariate, chandola2009anomaly,pena2013anomaly,tinawi2019machine,malhotra2015long,ahmad2017unsupervised,salinas2020deepar}.  

\subsubsection{Reconstruction-based method} In this AD approach, time series  data is reconstructed, and then by comparing the real (true) and 
reconstructed values, anomalies are detected \cite{ su2019robust,geiger2020tadgan,li2019mad,9669010}. As this approach must reconstruct the time series  to detect anomalies, it cannot be performed in real-time. 

Although both reconstruction and forecasting-based approaches can capture the anomaly points, they cannot provide confidence intervals of an NN model.  However, model uncertainty is essential for assessing how much to rely on the forecast produced by the model and it plays a critical role in  applications like AD. 
The uncertainty measure can be the variance or confidence intervals around the prediction made by the NN. In this paper, we use MC dropout to  capture the variance and to construct confidence interval for the NASA satellite telemetry dataset \cite{hundman2018detecting}. Specifically, by using MC dropout, we derive the  first- and second-order statistics of an NN consisting of the LSTM cells and the dense layers.  
These statistics are used to determine which points fall inside the confidence region and flag tentative points that are outside the region. In this paper, we describe how to use Bayesian LSTM for fault detection, identification, and recovery (FDIR)
for spacecrafts. We show the proposed MC-dropout approach outperforms the existing methods \cite{hundman2018detecting, ding2018multivariate, chandola2009anomaly,pena2013anomaly,tinawi2019machine,malhotra2015long,ahmad2017unsupervised,salinas2020deepar,su2019robust,geiger2020tadgan,li2019mad,9669010} and it 
has several advantages,  which will be discussed in detail in the next sections.

The rest of this paper is organized as follows: in Section II, we present the related works and lay out our contributions. In Section III, we discuss the proposed pre-processing method.   Bayesian LSTM methods for training and  post-processing AD filters are presented in Section IV.  Performance evaluation of the proposed AD method is presented in Section V followed by concluding  remarks in Section VI.  

\section{Related Works }
In this section, we first review the literature on different AD methods, compare the existing works from various perspectives, and then discuss this paper's contributions. 
\subsection{Literature Review}
 While forecasting-based methods suffer from sensitivity  to parameter selection and often require strong assumptions and extensive domain knowledge
about the data, they grant a significant amount of interest due to their low complexity. In this respect, some statistical-based models have been proposed, such as
Auto-regressive Integrated
Moving Average (ARIMA), which is an analytical model that learns the time series auto-correlation  for
future value predictions \cite{pena2013anomaly}. Furthermore, different sequential based deep neural network (NN) approaches such as LSTM, DeepAR, HTM 
are also forecasting-based methods \cite{malhotra2015long,salinas2020deepar,ahmad2017unsupervised}. In \cite{ahmad2017unsupervised}, the authors introduce hierarchical temporal memory (HTM) for AD in streaming data. The HTM encodes the current input to a hidden state and predicts the next
hidden state.  
  In \cite{malhotra2015long},  the performance of LSTM for detecting anomalies in the space shuttle dataset is discussed. In \cite{salinas2020deepar}, the  authors used
 DeepAR  which is based on autoregressive recurrent networks. In this approach, anomalies are computed as the regression errors which are the
distance between the median of the predicted and the  true
values.

 On the other hand, reconstruction-based methods aim to learn a model in order to capture the low-dimensional structure of the  data. Different from  the normal points, the anomalies lose more information when mapping to a lower dimensional space, so, as a result, they 
cannot be effectively
reconstructed. Thus, the points with high  reconstruction errors are more likely to be anomalies. In this respect, generative adversarial networks (GANs) for time series 
AD are proposed in \cite{geiger2020tadgan,9669010}. The OmniAnomaly method in \cite{su2019robust}
proposes a stochastic recurrent neural network (RNN), which captures
the normal patterns of multivariate time series  by modeling
data distribution through stochastic latent variables.  Furthermore, 
In \cite{hundman2018detecting}, the LSTM Auto-Encoder (AE) efficiency for detecting the  satellite anomalies is demonstrated. Recently, 
in \cite{choi2022multivariate}, the authors proposed a reconstruction-based AD method based on  a combination of  AE and convolutional NN  to capture the temporal correlations and spatial features of the multivariate time series. In \cite{jeong2022time}, an implicit neural representation-based AD approach is utilized.  In \cite{tuli2022tranad, yang2021improved},  a deep transformer NN which uses attention-based sequence encoders for AD is developed. More specifically, in \cite{yang2021improved} the prediction accuracy and AD are shown to be significantly improved  by using an attention-based mechanism. 

Nevertheless, both reconstruction and forecasting-based approaches are shown  promising results in AD,  they cannot provide a confidence region of the predicted points by NN. In the context of deep NN, the uncertainty region is defined as the likelihood interval for the true value that the  NN prediction lies within, and  the main approach to construct this interval is Bayesian Neural Network (BNN) \cite{loquercio_segu_2020}.  Using this uncertainty region and predicted value by NN, we can determine the anomaly points.  
As BNN does not suffer from over-fitting and always offers uncertainty estimation, it grants a significant amount of interest in many applications\cite{Gal2015}. However,
  finding the  posterior distribution  of BNN is 
 challenging and  computationally intractable. 
Therefore, the posterior distribution needs to be approximated with different techniques   \cite{NIPS2011_7eb3c8be,bui2016deep,mirikitani2008dynamic, Gal2015,9626568}.    Among various approximation approaches, the MC dropout has demonstrated important benefits, such as lower computational cost and higher precision, over other  methods \cite{Gal2015}.  The usage of MC dropout for RNN (i.e., LSTM) and the mathematical formulation can be found in \cite{gal2016theoretically}.  In \cite{chen2021imbalanced}, the authors proposed an AD technique based on an approximation of BNN. 
Furthermore, in  \cite{MOU2022103619,zgraggen2022uncertainty}, application  of BNN for AD is discussed. Moreover, in \cite{6823144}, the authors used the inference abilities and modeling characteristics of dynamic Bayesian networks in developing and implementing a innovative approach for FDIR of autonomous spacecraft. 
\subsection{Contributions}
In this subsection, we discuss  the contributions, advantages, and major differences of the proposed Bayesian method compared to other competitors.
  The key contributions of this paper are summarized as follows:  

 \begin{itemize}
     \item We propose an unsupervised Bayesian AD  method for time series data. In
particular, we use the MC dropout as a low-complexity approximation of BNN  in both the dense and LSTM layers of the NN for AD tasks. The proposed Bayesian method not only forecasts the data  accurately but also  provides a confidence interval as a byproduct. Note that,   our approximation of BNN is more precise than the method in  \cite{chen2021imbalanced}.  The MCD-BiLSTM-VAE method in \cite{chen2021imbalanced}, uses dropout only before the dense layers rather than the entire network composed of both the LSTM cells and the dense layers, and thus it only captures the uncertainty of the dense layers.  In contrast, we applied MC dropout at both of the LSTM and dense layers. Indeed, to get a precise approximation for BNN, the uncertainty of both LSTM and the dense layers should be acquired simultaneously \cite{gal2016theoretically,Gal2015}. Furthermore, another significant difference between our approach and \cite{chen2021imbalanced} in AD is that  the thresholding approach in \cite{chen2021imbalanced} is static, whereas we  propose a dynamic thresholding method. For the dataset under consideration, the dynamic thresholding method surpasses the approach of \cite{chen2021imbalanced} in terms of more accurate prediction and better classification results.
\item We propose a pre-processing method, which is based on an adaptive weighted average inside a time window with a variable length to smooth the time series. In our pre-processing method, unlike the standard pre-processing steps such as normalization and data cleaning, which are implemented in  \cite{chen2021imbalanced,hundman2018detecting,tinawi2019machine,li2019mad,geiger2020tadgan}, we first clean, normalize and scale the dataset, then implement an innovative  algorithm.  To the best of our knowledge, no prior work can be found that uses this method.  
 \item The proposed Bayesian method is a low complexity approach in terms of training time and throughput. In fact, mathematically, it can be shown that the dropout itself reduces the complexity in terms of the number of operations during both of the training and testing processes \cite{gao2016dropout}. Moreover,  comparing to the reconstruction-based methods such as \cite{chen2021imbalanced, geiger2020tadgan,hundman2018detecting},  
 our method is less computationally complex since it uses a smaller portion of the time series for predicting the next step. Indeed, at each time step, we use a number of data points equal to the order of the Markov model (which in practice will generally be one or two points) to estimate the next step and detect tentative anomalies. In contrast, reconstruction-based methods utilize all the time-series samples for AD. Consequently, they cannot be implemented in the real-time. 
\item Our proposed method prevents the model from overfitting, which is due to adding the regularization term to the loss function \cite{gao2016dropout}.   This is  different from the methods in \cite{hundman2018detecting,tinawi2019machine,li2019mad,geiger2020tadgan}, which might suffer from over-fitting. 
\item   We propose a post-processing algorithm which is based on the divergence of a succession of predicted data points from the corresponding confidence intervals.  This is different from   the  approaches in \cite{hundman2018detecting,chen2021imbalanced,geiger2020tadgan},  which suffer from a high number of false positives. The proposed post-processing approach is shown to lessen the number of false positives and, as a result, improves the accuracy of AD. 

 \end{itemize}

\section{Pre-processing for anomaly detection }
The raw telemetry data could not be directly ingested into our AD algorithm, and it should first 
go through  a pre-processing stage \cite{hundman2018detecting}.  An special  weighted moving average (WMA) filter is  adopted to improve the prediction accuracy by making the time series  smoother \cite{bifet2007learning}. In our terminology, the  high prediction accuracy is equivalent to low  mean square error (MSE) between the real values and the predicted values.  However, when there exist  anomaly points inside a window of the time series, the standard WMA may  not be applicable. In fact, due to  the averaging operation between the current  and the  adjacent points, there exists a high chance of missing an anomaly point.      
 Our motivation here is to devise a strategy to
 take the advantages of the WMA for improving the prediction accuracy without missing any anomaly point. In fact, we average among the consecutive data points within an  adaptive rolling window with a variable length. The idea lies in the  fact that when the difference between two consecutive measurements is high, there exists a high chance of having an outlier/anomaly point. Therefore, a narrower window length should be used, as we do not want the anomaly point to be missed before applying the anomaly detection algorithm.  Alternatively, when the difference between two consecutive points is small, we choose a wider window to make the time series  smoother.  
 Assume that 
${\bf u} \in \mathbb{R}^{n \times 1} $ and ${\bf{x}} \in \mathbb{R}^{n \times 1}$ are the raw data and the pre-processed data,  respectively. Let $l_w(k)$ be the length of the $k$th  window of our filter.  
   Assume that the processed data in the $k$th window is ${\bf{x}}(k)$, where $k \in {\cal{J}}, {\cal{J}}=\{1,...,n\}$.
      The smoothed data is given by:
\begin{equation}
{\bf{x}}(k) = \frac{1}{{\left\| \kappa  \right\|}}\sum\limits_{i =  - \frac{{{l_w(k)}}}{2}}^{\frac{{{l_w(k)}}}{2}} {{\alpha _i}{\bf{u}}(k - i)} ,
\end{equation}
where $\kappa  = {[ {\begin{array}{*{20}{c}}
{{\alpha _{  \frac{{-{l_w(k)}}}{2}}}}& \ldots &{{\alpha _{\frac{{{l_w(k)}}}{2}}}}
\end{array}}]^T}$, ${\left\| . \right\|}$  is the vector norm function, and $\alpha _i,  i \in {\cal{I}} = \left\{ {{\mkern 1mu} {\mkern 1mu} \frac{{ - {l_w(k)}}}{2},...,\frac{{{l_w(k)}}}{2}\,} \right\}$ are the smoothed filter coefficients.  In our approach, 
the smoothing coefficients $0 \le {\alpha _i} \le 1$, depend on the distance function ${\bf{d}}$, where ${\bf{d}}(k-i)=\left| {{{{\bf{u}}(k-i-1) - {\bf{u}}(k -i )}}} \right|$. The distance function is the absolute difference between  two successive data points inside the $k$th window.  In this pre-processing algorithm, we start with  
$l_w(0)=2$ and ${\alpha _0=1}$. Let the distance threshold value be ${d_{th}^{(k)}} = m^{(k)} + 2\sigma^{(k)}$, where $m^{(k)}$ and $\sigma^{(k)}$ are the mean and standard deviation (s.t.d.) of the data inside the $k$th window.   We choose the threshold value   $d_{th}^{(k)} = m^{(k)} + 2\sigma^{(k)}$ to take into account both the mean and variance inside each window. However,  this threshold can be further improved by examining the exact distribution of different time series  and choosing the optimal value for the threshold. It remains an interesting avenue for our future studies.

For the $i$th  data in the $k$th window, if  ${\bf{d}}(k-i)<d_{th}^{(k)}$ (i.e., we do not see a sudden change), we increase the window length and set  ${\alpha _i} = 1, \,\, \,\,i \in \cal{I}$. Otherwise, if  ${\bf{d}}(k-i)\ge d_{th}^{(k)}$, we have 
\begin{equation}
    {\alpha _i} = \left\{ \begin{array}{l}
0\,\,\,\,\,\,\,\,\,\,i \in {\cal{I}} - \{ 0\} \,\,\,\,\,\,\\
1\,\,\,\,\,\,\,\,\,\,i = 0,\,\,\,\,\,\,\,\,
\end{array} \right. 
\end{equation}
 then we increment $k$, and proceed to another window, i.e., $k = k+1$.  The pre-processing procedure is shown in Algorithm 1.
\begin{algorithm}

\hspace*{\algorithmicindent} \textbf{Input:} ${\bf{u}}(k)$. \\
\hspace*{\algorithmicindent} \textbf{Output:} ${\bf{x}}(k)$.
\begin{algorithmic}[1]
\caption{Pre-processing algorithm}
\State   {Initialization: $l_w(0)=2$, ${\alpha _0=1}$, $k=0$}.
\While { $k \le n$ }
\State{Calculate the mean $m^{(k)}$ and s.t.d. $\sigma^{(k)}$ of the values inside the $k$th windows with length $l_w(k)$.}
\State{Select the $k$th threshold as $d_{th}^{(k)} = m^{(k)} + 2\sigma^{(k)}$}
\For{$i \in {\cal{I}}$} 
\If{ ${{\bf{d}}(k-i)}<d_{th}^{(k)}$.}
\State{${\alpha _{i}}=1$}
\State{$l_w(k)=l_w(k)+1$}
\ElsIf{ ${{\bf{d}}(k-i)}>d_{th}^{(k)}$}
\State{Break}
\EndIf
\EndFor
\State{ Calculate ${\bf{x}}(k) = \frac{1}{{\left\| \kappa  \right\|}}\sum\limits_{i =  - \frac{{{l_w(k)}}}{2}}^{\frac{{{l_w(k)}}}{2}} {{\alpha _i}{\bf{u}}(k - i)}$.}
\State{Go to the next window, i.e., $k=k+1$.}
\EndWhile
\end{algorithmic}
\end{algorithm}
\section{Bayesian LSTM method for prediction and uncertainty estimation}
\begin{figure}[ht!]
    \includegraphics[width=3.2in]{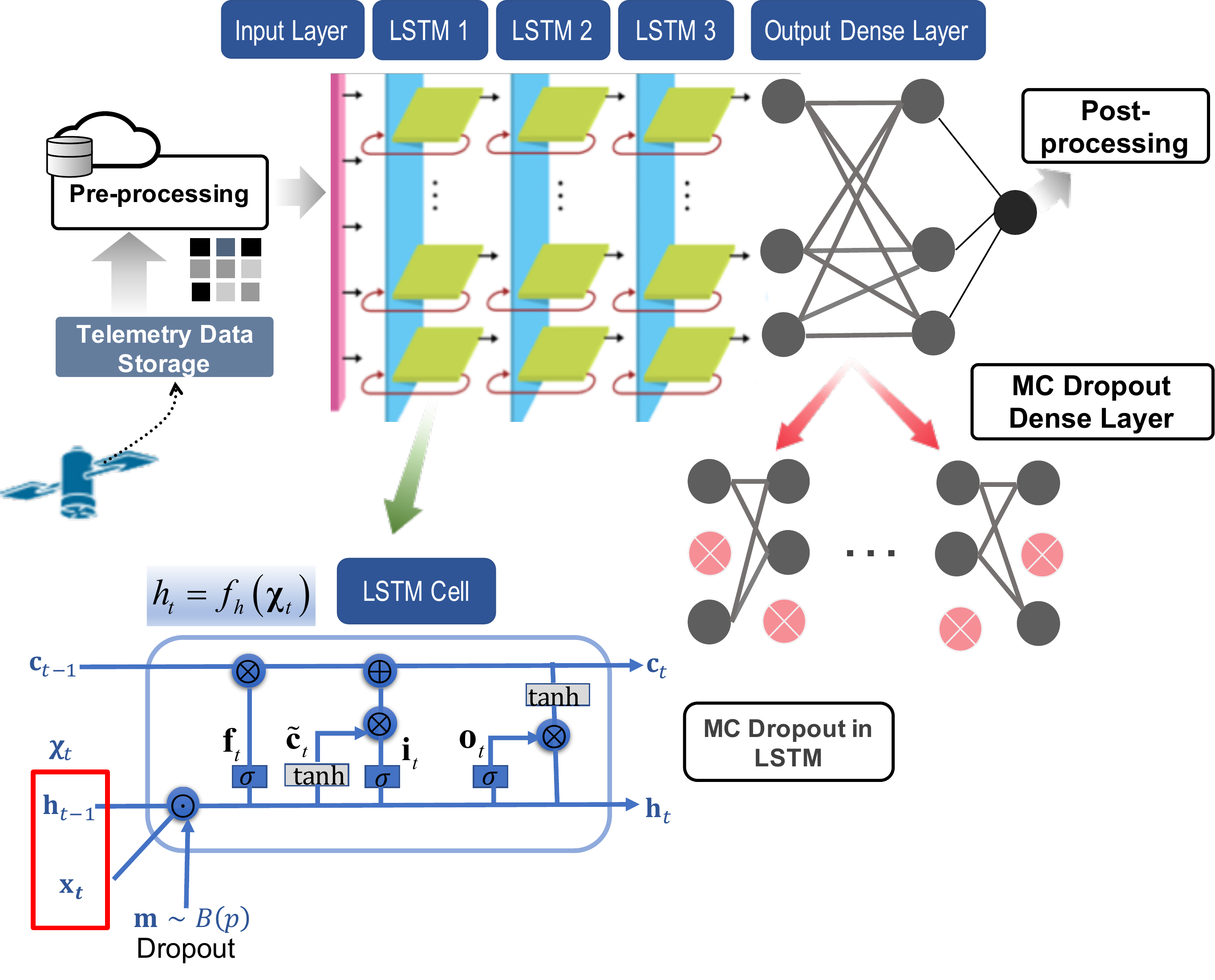}
    \caption{An illustration of the proposed system model using Bayesian LSTM}
    \label{Fig1}
\end{figure}

Here, we discuss a Bayesian LSTM approach  for predicting the mean of the model and its associated variance (uncertainty level). 
 Fig.~\ref{Fig1} shows our proposed NN model and  the underlying Bayesian  approach, where we use several LSTM cells followed by the dense layers. In this section,  we adopt the MC dropout for  approximation of the BNN model in both of the LSTM and dense layers separately. 
\subsection{MC dropout as an approximation of BNN}
We assume an NN with $L$ LSTM and $D$ dense layers.   
A batch of $T$ observations is passed through the input layer.  By assuming $T_D$ as the length of the $D$th dense layer, we  denote ${\bf x} \in \mathbb{R}^{T}$ and ${\bf y} \in \mathbb{R}^{T_D}$  as the input and output of NN, respectively. 
Assume that ${\boldsymbol \omega} $ with prior distribution $p(\bf{\boldsymbol \omega})$ represents a collection of the NN  parameters. The probability of the output vector given the data for  an associated  input vector ${\bf{x}}$ is:
 
\begin{equation}
    \label{eq6}
    p\left( {{\bf{y}}\left| {{\bf{x}},{\bf{D}}} \right.} \right) = {{\rm{E}}_{p\left( {{\boldsymbol{\omega }}\left| {\bf{D}} \right.} \right)}}\left( {p\left( {{\bf{y}}\left| {{\bf{x}},{\boldsymbol{\omega }}} \right.} \right)} \right), 
\end{equation}
where ${\bf{D}} = \{ \left( {{x_t},{y_t}} \right)\,\,\forall t \in \{ 1,...,T\} \} $ and $p({\boldsymbol{\omega}}|{\bf D})$ are a batch of data, and the posterior distribution on
weights, respectively. The expectation under $p({\boldsymbol{\omega}}|{\bf D})$  is equivalent to using an ensemble of an infinite number of models, which is indeed computationally intractable. 

 One feasible solution is to approximate the posterior distribution over the model parameters $p\left({\boldsymbol \omega \left| {\bf D} \right.} \right)$ with a simpler distribution  $q(\boldsymbol \omega)$.  
 A metric of similarity (or distance) between two probability distribution functions (PDF) is Kullback-Leibler (KL) divergence. Here, we  minimize the
${\rm{KL}}(q({\boldsymbol{\omega }})\left\| {p\left( {{\boldsymbol{\omega }}\left| {\bf{D}} \right.} \right)} \right)$  which is the KL distance between ${p\left( {{\boldsymbol{\omega }}\left| {\bf{D}} \right.} \right)}$ and $q({\boldsymbol{\omega }})$ \cite{Gal2015}. More specifically, we solve the following  optimization problem:

    \begin{equation}
        \mathop{{\rm{min}}}\limits_{q({\bf{\omega }})} \,\,\,\,\,{\rm{   KL}}(q({\bf{\omega }})\left\| {p\left( {{\bf{\omega }}\left| {\bf{D}} \right.} \right)} \right.),    
    \end{equation}
which is equal to  minimizing the following:
\begin{equation}
     \label{awe}
  \!\mathop {{\rm{min}}}\limits_{q({\bf{\omega }})} \,\, - \!\int {q({\bf{\omega }})\log p({\bf{y}}\left| {\bf{x}} \right.,{\bf{\omega }})d{\bf{\omega }} + {\rm{KL}}(q({\bf{\omega }})\|p({\bf{\omega }}))}.   
\end{equation}
where $p({\boldsymbol \omega})$  stands for the prior distribution of the NN parameters. 
 As seen in (\ref{awe}), the objective function is composed of two parts. The first component is the NN loss function and the second one stands for the regularization effect, which prevents the model from being over-fitted. Minimizing the second component is equivalent to finding a $q(\omega)$ close to the prior, which essentially avoids over-fitting. The first component, on the other hand, can be rewritten with the MC sampling over $\boldsymbol \omega$ with a
single sample as the following: 
\begin{equation}
\label{eqq7}
 \,\,-\frac{1}{T}\,\, \sum\limits_{n = 1}^T \int{\,\log p({{\bf{y}}_n}|{f^{\boldsymbol{\omega}}({\bf{x}}_n}}))d{\boldsymbol{\omega}}.
\end{equation}

In the next subsection, we discuss the detailed structure of the function ${f^{\boldsymbol{\omega}}({\bf{x}}_n})$ in (\ref{eqq7}) using different layers of LSTM followed by dense layers. 
\subsection{Application of MC dropout for 
approximation of the posterior distribution of NN parameters}
As shown in Fig.~\ref{Fig1}, a simple LSTM unit contains input, output, and additional control gates. The following formulation briefly shows the principle of the LSTM\cite{hochreiter1997long}:

\begin{equation}
    \begin{array}{l}
{{\bf{i}}_t} = \sigma \left( {{{\bf{W}}_i}{{\bf{\chi }}_t} + {{\bf{b}}_i}} \right),\,\,\,\,\,{{\bf{f}}_t} = \sigma \left( {{{\bf{W}}_f}{{\bf{\chi }}_t} + {{\bf{b}}_f}} \right),\\
{{\bf{o}}_t} = \sigma \left( {{{\bf{W}}_o}{{\bf{\chi }}_t} + {{\bf{b}}_o}} \right), \,\,\,\,\,
 {{{\bf{\tilde c}}}_t} = \tanh \left( {{{\bf{W}}_{\tilde c}}{{\boldsymbol \chi} _t}} +{{\bf b}_c}\right),\\
 {{\bf{c}}_t} = {{\bf{f}}_t} \odot {{\bf{c}}_{t - 1}} + {{\bf{i}}_t} \odot {{{\bf{\tilde c}}}_t}, \,\,\,\,\,
 {{\bf{h}}_t} = {{\bf{o}}_t} \odot {\rm{tanh}}({{\bf{c}}_t}),
\end{array}
\end{equation}
where ${{\boldsymbol \chi} _t} = {\left[ {\begin{array}{*{20}{c}}
{{{\bf{x}}_t}}&{{{\bf{h}}_{t - 1}}}
\end{array}} \right]^T}$ and ${\bf x}_t$ represents the input data, $\sigma$ is the logistic sigmoid function and $\odot$ is the element-wise product. Also,  ${{\bf{c}}_t}$ represents the state value of the LSTM
unit (while ${{{\bf{\tilde c}}}_t}$ is the candidate state value),  ${{\bf{i}}_t}$  stands for the state value
of the input gate, $ {{\bf{f}}_t} $ represents the state value of the forget gate,  ${{\bf{o}}_t}$ is the  state value of the output gate, $ {{\bf{h}}_t} $ is the output
of the LSTM unit, and $t$ is the current time step. ${\bf W}$ and ${\bf b}$  are the corresponding weight and bias
parameters at each of the aforementioned gates in (6).
According to the aforementioned formulation of LSTM, the following mapping function can be concluded (omitting the bias parameter): 
\begin{equation}
{{\bf{h}}_t} = {f_h}({\boldsymbol{\chi _t}}).
\end{equation}

By applying a dropout at the input and hidden layers, we have \cite{gal2016theoretically}:
\begin{align}
\label{eq12}
 {{\bf{i}}_t} &= \sigma \left( {{{\bf{W}}_i}({{\boldsymbol \chi} _t} \odot {\bf{m}})}+{{\bf b}_i} \right),\\
 {{\bf{f}}_t} &= \sigma \left( {{{\bf{W}}_f}({{\boldsymbol \chi} _t} \odot {\bf{m}})+{{\bf b}_f}} \right),\\
 {{\bf{o}}_t} &= \sigma \left( {{{\bf{W}}_o}({{\boldsymbol \chi} _t}\odot {\bf m})+{{\bf b}_o}} \right),\\
 {{{\bf{\tilde c}}}_t} &= \tanh \left( {{{\bf{W}}_{\tilde c}}({{\boldsymbol \chi} _t} \odot {\bf{m}})+{{\bf b}_c}} \right),\\
 {{\bf{c}}_t} &= {{\bf{f}}_t} \odot {{\bf{c}}_{t - 1}} + {{\bf{i}}_t} \odot {{{\bf{\tilde c}}}_t},
\end{align}
where ${\bf{m}} \sim {\cal{B}}(p)$  is a random mask vector where each row has a  Bernoulli distribution with the parameter $p$ and it is repeated at all the time steps.  
We use the MC dropout in the dense layers by dropping the neurons randomly according to the  Bernoulli distribution during the testing phase. For the $k$th dense layer, we have \cite{Gal2015}:  
\begin{align}
{{\bf{W}}_k} &= {\rm{diag}}({[{z_{k,j}}]_{j = 1}^{{K_i}}}){{\bf{M}}_k},\\ \nonumber
{z_{k,j}} &\sim {\cal{B}}\left({{p_k}} \right)\,\,\,\,\,\forall k \in \{1,...,D\}\,,\,j  \in \{1,...,{K_{k-1}}\},
\end{align}
where $p_k$ and matrix ${{\bf M} _k}$ of dimensions $K_k \times K_{k-1}$ are the variational
parameters. The binary variable ${z_{k,j}}= 0$ corresponds
to the unit $j$ in layer $k-1$ being dropped out as an input to the
 $k$ layer. 
 By assuming ${\bf b}_k$ as a bias at the $k$th dense layer,  the weight matrix (NN parameter) can be considered as ${\boldsymbol{\omega}}= [{\bf{W}}_k,{\bf{W}}_i, {\bf{W}}_f, {\bf{W}}_o, {\bf{W}}_{\tilde c},{\bf b}_k,{\bf b}_i,{\bf b}_f,{\bf b}_o,{\bf b}_c ]^T$.
Therefore, (\ref{eqq7}) is rewritten as
\begin{equation}
 \resizebox{.9\hsize}{!}{$\sum\limits_{i = 1}^N {\log ( {p({y_i}|\sqrt {\frac{1}{{{K_D}}}} {{\bf{\hat W}}_D}\sigma (...\sqrt {\frac{1}{{{K_2}}}} {{\bf{\hat W}}_2}\sigma( {{{\bf{\hat W}}_1}({\boldsymbol{\psi _T}}) + {{\bf{b}}_1}})...})))}$},
\end{equation}
where ${\boldsymbol{\psi }}_T \buildrel \Delta \over = [{{\bf{x}}_{T,i}},f_h^{\hat {\boldsymbol{\omega}} }(...f_h^{\hat {\boldsymbol{\omega}}}([{{\bf{x}}_{0,i}},{{\bf{h}}_0}])]$, and  ${\hat {\boldsymbol{\omega}}}  \sim q\left({\boldsymbol \omega}  \right)$. 
 For each sequence ${\bf{x}}_i$, we sample a new realization ${\boldsymbol{\hat \omega}}= [{\bf{\hat W}}_k,{\bf{\hat W}}_i, {\bf{ \hat W}}_f, {\bf{ \hat W}}_o, {\bf{\hat W}}_{\tilde c},{\bf \hat b}_k,{\bf \hat b}_i,{\bf \hat b}_f,{\bf \hat b}_o,{\bf \hat b}_c ]^T$.
 Each symbol in the sequence ${\bf x}_i = [{\bf x}_{i,1}, ...,{\bf x}_{i,T} ]^T$ is passed through the function $f_h^{\hat {\boldsymbol{\omega}} }$ with the same
weight realizations ${\hat {\boldsymbol{\omega}}}$ used at the time step $t \le T$. In the dense layers, and LSTM layers, the  MC dropout is performed during both of the training and testing phases.   To summarize, in MC dropout, we set the input of each neuron (or LSTM  cell)  independently to zero with probability $p$, which means
running the network several times with different random seeds. Algorithm 2 shows the different steps of the MC dropout approach.   
\begin{algorithm}
\begin{algorithmic}
\caption{MC dropout algorithm}
\State Repeat: 
    \State  Sample ${z_{i,j}} \sim {\cal{B}}\left({{p_i}} \right), \,\,\,\,\, i \in \{1,...,D\}\,,\,j \in \{1,...,{K_{i - 1}}\}$, and ${\bf{m}} \sim {\cal{B}}\left({{p_k}} \right)\,\,\,\,\,\forall k \in \{1,...,L\} $ and set (12), (13), (14), (15), (16),  (17), respectively, and find\\  
${\boldsymbol{\hat \omega}}= [{\bf{\hat W}}_k,{\bf{\hat W}}_i, {\bf{ \hat W}}_f, {\bf{ \hat W}}_o, {\bf{\hat W}}_{\tilde c},{\bf \hat b}_k,{\bf \hat b}_i,{\bf \hat b}_f,{\bf \hat b}_o,{\bf \hat b}_c ]$ 
where ${\boldsymbol{\hat \omega}}\sim q({\boldsymbol{ \omega}})$.
    \State Minimize (one step):
\begin{equation}
-\int  q({ \boldsymbol  \omega }) \log \left({p\left({{\bf{y}}\left| {{\bf{x}},} \right. {  {\boldsymbol \omega}} } \right)} \right) \nonumber
+{\rm  KL}\left({{q }\left(\boldsymbol  \omega  \right), {p\left(\boldsymbol \omega  \right)}} \right). 
\end{equation}
\end{algorithmic}
\end{algorithm}

Assume that we have $l$ sets of realization of the NN after applying the MC dropout, as shown in Fig. \ref{Fig1}. 
As derived in  \cite{Gal2015}, the approximated predictive distribution is given by the following:
\begin{equation}
q({{\bf{ y}}}\left| {{{\bf{ x}}}} \right.){\rm{ }} = {\int} {p({{\bf{ y}}} | {{{\bf{ x}}}},{\bf{\boldsymbol \omega }})q({\bf{\boldsymbol \omega }})d{\bf{\boldsymbol \omega }}} 
\end{equation}
The first two moments are derived as\cite{Gal2015}:
\begin{equation}
{{\rm E}_{q({\bf{y}}  \left| {{\bf{ x}} } \right.)}}({{\bf{y}}}) \approx \frac{1}{l}\sum\limits_{t = 1}^l {{{\rm E({\bf{  y}}}}|{{\bf{ x}}},{\bf{{ \boldsymbol \omega}}}^t})
\end{equation}
and 
\begin{equation}
{{\rm E}_{q({\bf{y}}  \left| {{\bf{ x}} } \right.)}}({ {{{\bf{ y}}}}^T} {{{\bf{ y}}}}) \approx 
  \frac{1}{l}\sum\limits_{t = 1}^l {{{{\rm E(\bf{y}}|}}{{{{\bf{ x}}},{\bf{ {\boldsymbol \omega}}}^t)}^T}} {{\rm E({\bf{ y}}|}}{{\bf{ x}}},{\bf{ {\boldsymbol \omega}}}^t)
\end{equation}
The model’s predictive variance is
\begin{equation}
 {\rm{Var}}{_{q({\bf{y}}  \left| {{\bf{ x}} } \right.)} }({\bf{y}}) \!=\! {{\rm{E}}_{q({\bf{y}}  \left| {{\bf{ x}} } \right.)} }({{\bf{y}}^T}{\bf{y}}) \!-\! {{\rm{E}}_{q({\bf{y}}  \left| {{\bf{ x}} } \right.)} }{({\bf{y}})^T}{{\rm{E}}_{q({\bf{y}}  \left| {{\bf{ x}} } \right.)} }({\bf{y}})   
\end{equation}
 
\subsection{Post-processing of AD}
To reduce the number of false positives, we propose a post-processing algorithm, which is based on the deviation of a sequence of predicted data points from the confidence region. More specifically, if a predicted data point is outside the uncertainty region of the MC dropout, then it is considered as  a tentative anomaly point. In our specific dataset, anomaly points do not occur as a single point; instead, they emerge as a sequence (burst) of dependent points in a particular part of the time series.

It is important to determine a time when the anomaly starts. Let $N_{max}$ be a range of consecutive data points in time series. In our approach, if $0.8{N_{max}}$ of the consecutive  points are outside the confidence region simultaneously, then we consider the first index of the sequence as the starting point of the anomaly sequence. For this specific dataset, this post-processing approach can significantly reduce the number of false positives. Choosing a suitable value for $N_{max}$ depends on the maximum allowable delay. We can improve the accuracy if we wait for more sample points to announce the anomaly; however, successful alert with a huge delay is not useful. Therefore, if an AD algorithm raises an alert fast enough (i.e., before a maximum allowable delay) the whole anomaly fragment is considered to be detected successfully. We will further illustrate this method in numerical results. 



\section{Numerical Results}
First, we describe the methodology for evaluating AD, then summarize the specific dataset under consideration, followed by the performance evaluation. The source code is available at \url{https://bit.ly/3zf1Ipb}.  

\subsection{Methodology}
We assume an NN with 3 LSTM and 2 dense layers, where the MC dropout rate is assumed to be 0.2.  
We adopt PyTorch, an open-source framework for machine learning, to implement our proposed approach.  
We use the dataset in \cite{hundman2018detecting} to compare and evaluate our method with the following approaches: i) dynamic thresholding  approach with LSTM Auto-Encoder (AE)  \cite{hundman2018detecting}; ii) TadGan approach \cite{geiger2020tadgan}; iii) MadGan method \cite{li2019mad}; iv) Arima method  \cite{pena2013anomaly};  v) LSTM method \cite{tinawi2019machine}; and vi) A modified version of MCD-BiLSTM-VAE method in  \cite{chen2021imbalanced} after applying the  post-processing filter which has been proposed in Section IV.C. We will use the following metrics to evaluate the performance of various 
models:   mean squared error (MSE), $F_1$ score, accuracy, recall, and precision. Moreover, we will compare the computational complexity of different methods. These metrics have been widely
used in the literature and are discussed  in detail in \cite{mohri2018foundations}. 
\subsection{NASA dataset}

We use the satellite telemetry dataset, which was originally collected by NASA \cite{hundman2018detecting}. The dataset comes from two spacecrafts: the Soil Moisture Active Passive (SMAP) and the Curiosity Rover on Mars (MSL).   
There are 82 signals available in the NASA dataset, which are summarized  in Table \ref{T1}. We found  54 of the 82 signals (multivariate time series) are continuous, and the remaining  are discrete. There are a total of $n$ rows corresponding to the number of
timestamps or readings taken for that signal ($n$ differs in different signals). Each sample of the  data consists of one column of telemetry signal and several columns of commands. The command features are encrypted and can barely be  used. Furthermore, we also found that the correlation between the command features and telemetry signals is negligible. We   consider only the time series  sequences from the telemetry signals in our evaluation. Therefore, the input signal to our model 
is an $n\times 1$ matrix.
\begin{table}[ht!]
\centering
    \caption{NASA telemetry dataset  summary}
   \scalebox{0.9}{ \begin{tabular}{|c|c|c|c|c|}
    \hline
   Dataset&Signals &   Anomaly Sequences &Data Points & Anomaly Points\\ \hline
      SMAP  &55 & 67 & 562800&54696 \\ \hline
      MSL &27& 36&  132046 & 7766\\
      \hline
    \end{tabular}
}    
   \label{T1}
\end{table}

\begin{figure}[ht!]
    \centering
    \includegraphics[width=3.4in]{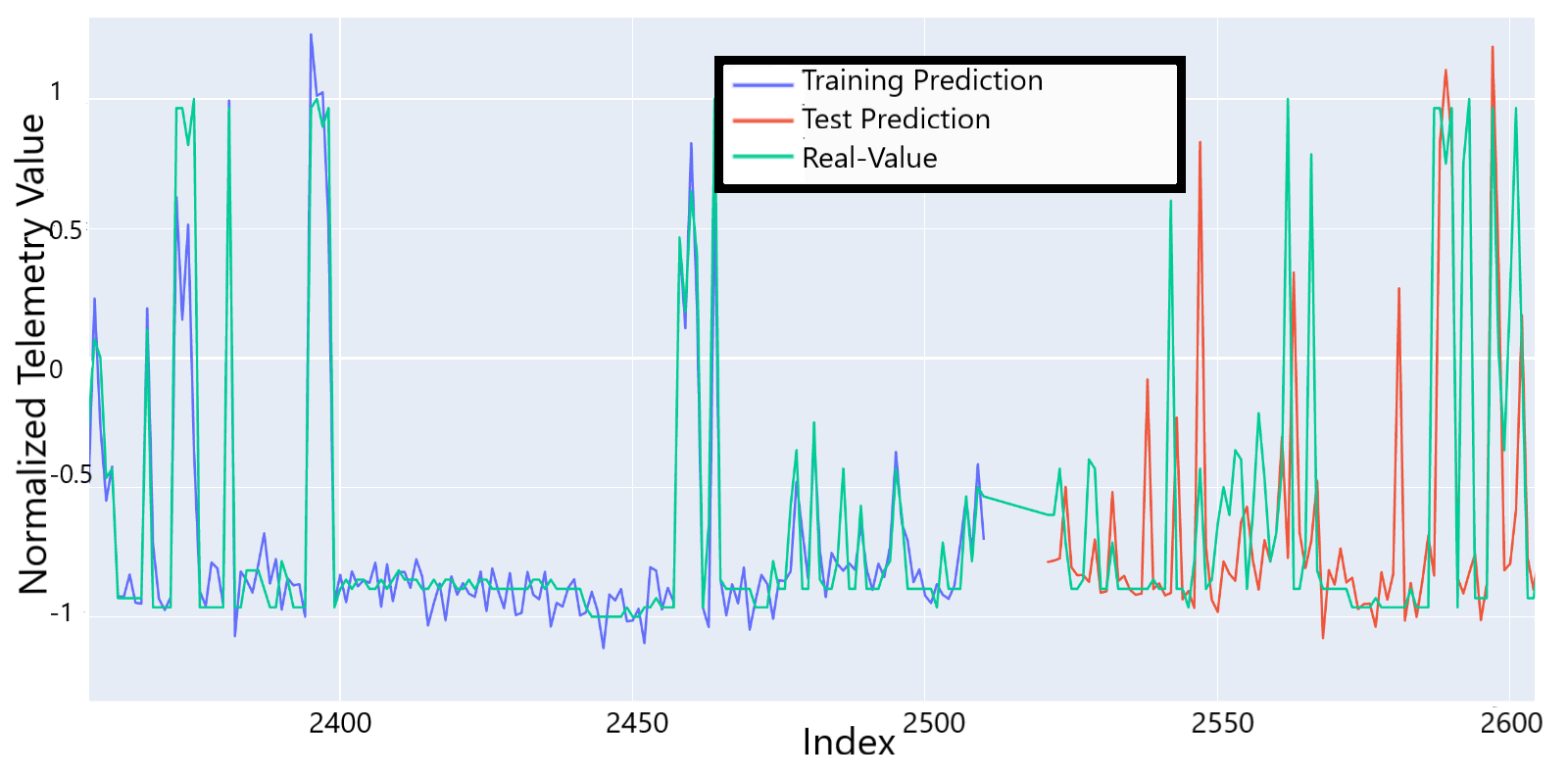}
    \caption{Examples of real values and predictions in training and testing modes  for `F-7' signal }
    \label{Fig2}
\end{figure}

\begin{figure}[ht!]
    \centering
    \includegraphics[width=3.4in]{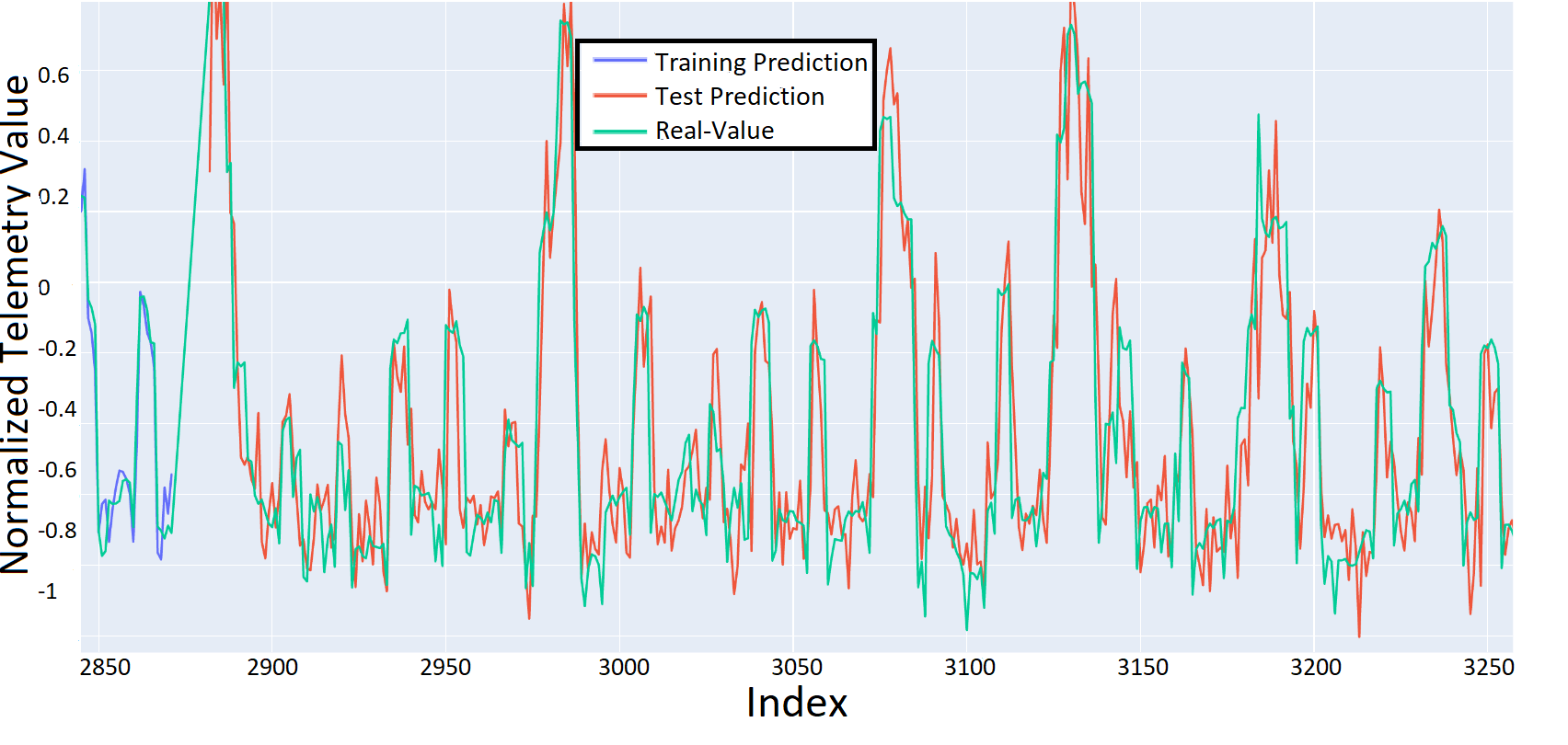}
    \caption{Examples of real values and predictions in training and testing modes   for `P-1' signal }
     \label{Fig3}
\end{figure}

\subsection{Evaluation}
\begin{figure}[ht!]
    \centering
    \includegraphics[width=3.4 in]{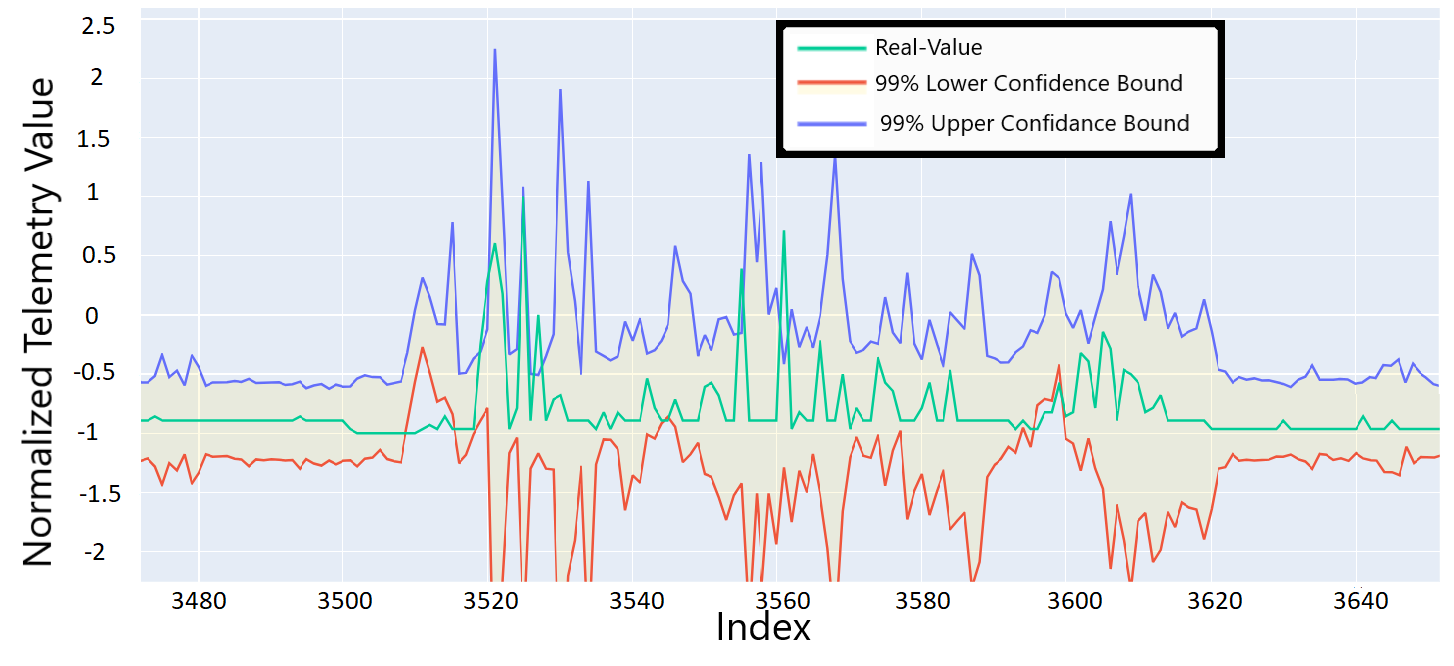}
    \caption{Uncertainty region using MC dropout for `F-7' signal}
    \label{Fig4}
\end{figure}

\begin{figure}[ht!]
    \centering
    \includegraphics[width=3.4 in]{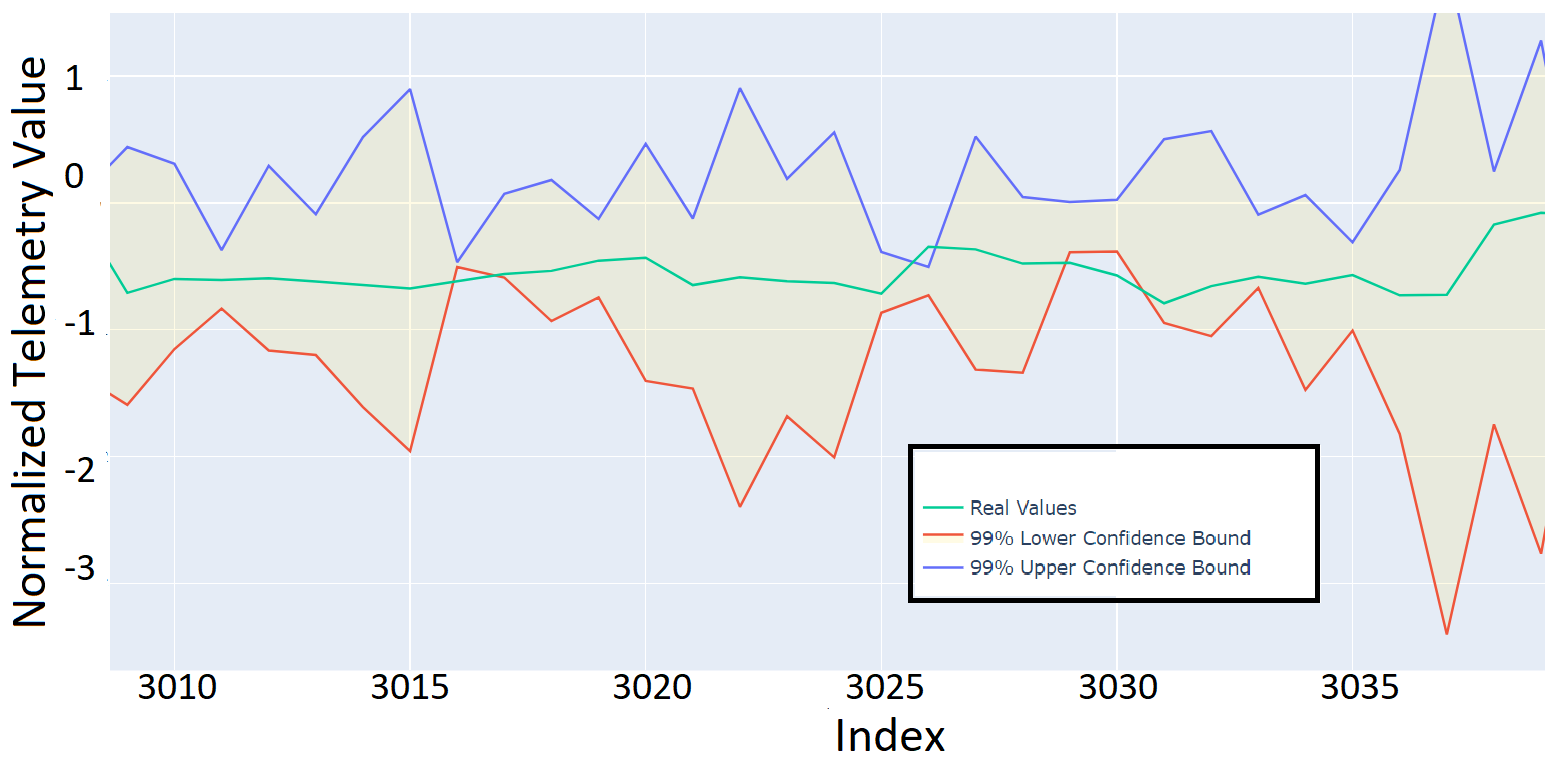}
    \caption{Uncertainty region using MC dropout for `P-1' signal}
    \label{Fig5}
\end{figure}

In this subsection, we evaluate and compare the proposed model with those of other studies. The comparison results  between the predicted and real values in the training and testing phases for one  of  the MSL satellite telemetry signals i.e., `F-7',  and another  SMAP satellite signal, i.e., `P-1'  are shown in Fig. \ref{Fig2} and Fig.~\ref{Fig3}, respectively.   By analysing these two figures, we notice that our Bayesian LSTM approach performs better on the SMAP data in the prediction phase.  
 To show the uncertainty region by our Bayesian approach, Fig. \ref{Fig4} and Fig. \ref{Fig5} are depicted. These  figures show the MC dropout bounds and corresponding predicted values for `F-7' from MSL dataset and `P-1' from SMAP dataset, respectively. We found that, on average, for both datasets,  $84\%$ of the predicted values are inside the  uncertainty bounds. Furthermore, we use the following  evaluation measures to assess our method: 

\subsubsection{Mean Squared Error}
 MSE, which is used as a metric for comparing forecasting-based methods, assesses the average squared difference between the real and predicted values. It is defined as 
${\rm{MSE}} = \frac{1}{N}\sum\limits_{i = 1}^N {{{\left( {{{\bf{y}}_i} - {{{\bf{\hat y}}}_i}} \right)}^2}} $
, where ${\bf{\hat y}}_i$ is the predicted value at the $i$th timestamp. The MSE results for the two datasets for our approach, the Arima method \cite{pena2013anomaly}, and the LSTM method  \cite{tinawi2019machine} can be found in Table \ref{T2}. The MSE can be used just for forecasting-based methods, and for the reconstruction-based methods such as TadGan and MadGan, LSTM AE, it cannot be calculated.   Indeed, what they  calculated is the quantity of error for reconstructing the time series  and it is different from the concept of MSE considered in the forecasting-based approaches. Due to the fundamental difference in definitions, we  have not compare the mentioned methods with our approach. 
\begin{table}[!ht]
\centering
   \caption{MSE comparison of forecasting-based approaches}
 \begin{tabular}{|c|c|c|c|c}
    \hline
   Baseline & Bayesian LSTM &  Arima \cite{pena2013anomaly} & LSTM \cite{tinawi2019machine} \\ \hline
      SMAP  &{0.05}& {0.61}&{0.07}\\\hline
      MSL &{0.21}&  0.83&{0.29}\\
      \hline
    \end{tabular}
    \label{T2}
\end{table}

\subsubsection{Recall}
This is often used as a sensitivity metric. It is the proportion of relevant instances that were retrieved. i.e.,  $\theta = \frac{{TP}}{{TP + FN}}$. 
where $TP$, $FP$, $FN$, and $TN$  stand for true positive, false positive, false negative, and true negative, in the confusion matrix, respectively.

\subsubsection{Precision}
It is the fraction of relevant instances among the retrieved instances, i.e., $\xi  = \frac{{TP}}{{TP + FP}}$.
\subsubsection{Accuracy}
It is one of the important classification performance  measures. Based on the definition, it is 
the proportion of correct predictions (both true positives and negatives) among the total number of explored cases, i.e.,
 ${\rm{Accuracy}} = \frac{{TP + TN}}{{TP + TN + FP + FN}}$.
 
\subsubsection{$F_{1}$ score} 
 This score  is defined as the harmonic mean of precision and recall, i.e.,
${F_1} = {\left( {\frac{{{\xi ^{ - 1}} + {\theta ^{ - 1}}}}{2}} \right)^{ - 1}}   $. 

\begin{figure}[ht!]
    \centering
    \includegraphics[width=3.5in]{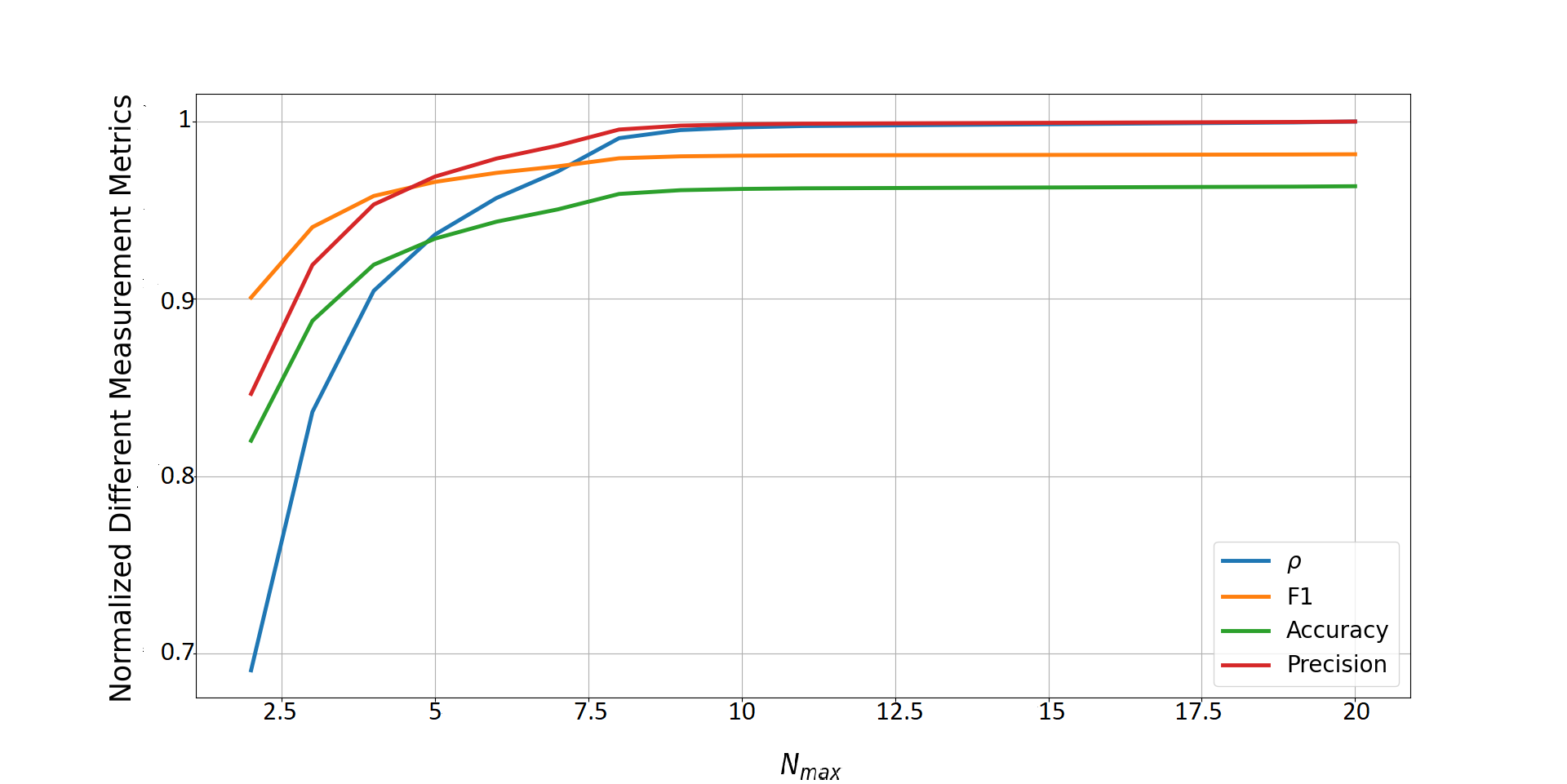}
    \caption{The normalized measurement scores v.s. $N_{max}$ for signal `P-1' to show the post-processing effect.} 
  \label{Fig6}
\end{figure}

For finding the best value of $N_{max}$ in post-processing,    
we perform a grid search among different values of $N_{max}$ for each signal independently and pick the optimal value in such a way that the  dissimilarity between the predicted labels and the known labels is minimized. The value of $N_{max}$ depends on the maximum allowable delay, and selecting a higher value for $N_{max}$, causes a smaller value for $FP$. Alternatively, choosing a high value for  $N_{max}$ (e.g., 100)  leads to a high value for $FN$. Therefore, finding the optimal value for this parameter will boost the AD overall  performance.    
 Assume a new metric
$\rho_n=T{P_n} + T{N_n} - F{P_n} - F{N_n}$. We choose ${N_{\max }}$ in such a way that $\rho_{n}$ gets maximized within a reasonable range of $n$.  
Fig. \ref{Fig6} shows the normalized value of different measurement criteria versus $N_{max}$ for signal `P-1'. This figure shows that after a specific amount of $N_{max}$, the value of $\rho_n$ (and other evaluation criteria) does not change. For this specific signal, the optimal value for $N_{max}$ is $N^{opt}_{max}=8$. 
Also, we found that the anomalies cannot be detected using some signals (e.g., ``M6'' in the MSL dataset has a value of -1 in all training and testing phases, which makes it impossible to be used in uni-variate AD cases). We have excluded these types of signals (i.e., ``M6, E3, A1, D1, D3,
D4, G1, D5, D11, G6, R1, A6, F3, M2, P10, M3, 
D16, P15, P11, P14'') in our comparison results.

\begin{table}[!ht]
\centering
   \caption{F1 score comparison}
   \scalebox{0.52}{ \begin{tabular}{|c|c|c|c|c|c|c|c|}
    \hline
   Baseline & Bayesian LSTM &  TadGan \cite{geiger2020tadgan} &Arima \cite{pena2013anomaly} &LSTM AE \cite{hundman2018detecting}& MadGan\cite{li2019mad}&LSTM \cite{tinawi2019machine}& MCD-BiLSTM-VAE\cite{chen2021imbalanced}\\ \hline
      SMAP  &0.84&{ 0.66}& 0.42 & 0.69 &0.12&0.62& 0.67\\ \hline
      MSL & {0.74}& {0.55} & 0.49& 0.55 & 0.11& 0.48 &{0.64}\\
      \hline
    \end{tabular}
}
   \label{T3}
\end{table}

\begin{table}[!ht]
\centering
\caption{Comparison of learnable parameters, training time, and inference throughput}
   \scalebox{0.7}{ \begin{tabular}{|c|c|c|c|}
    \hline
   Methods & Parameters &  Training Time (per epoch) & Throughput \\ \hline
   Bayesian LSTM &$229409$& {1.48} sec &  $0.57 \times 10^3$ predictions/sec \\
      \hline
      TadGan \cite{geiger2020tadgan} &$447189$& {$78$ sec} & {$0.27\times 10^2$} predictions/sec \\ \hline
      Arima \cite{pena2013anomaly} &{$17$}& 1.1 sec & {$2.1\times 10^3$} predictions/sec \\
      \hline
     LSTM AE \cite{hundman2018detecting} &$86250$&  1.84  sec & $1.16 \times 10^3$ predictions/sec \\ \hline
          LSTM \cite{tinawi2019machine} &229409& {1.48} sec & {$1.1\times 10^3$} predictions/sec \\ \hline
                 MCD-BiLSTM-VAE\cite{chen2021imbalanced} &580993&  6.96  sec & {$0.032 \times 10^3$} predictions/sec \\ \hline
    \end{tabular}}
   \label{T4}
\end{table}

\begin{table}
\centering
\caption{Comparison of accuracy, precision, and recall}
\label{tableV}
\scalebox{0.70}{
\centering
\begin{tabular}{|c|c|c|c||c|c|c|}
\hline
\multicolumn{1}{|c|}{\bfseries Methods} &
\multicolumn{3}{c||}{\bfseries SMAP}&
\multicolumn{3}{c|}{\bfseries MSL}\\
\hhline{|~|---||---|}
{} & Accuracy &  Precision & Recall & Accuracy &  Precision & Recall\\
\hline
{Bayesian LSTM} & {\textbf{0.75}} & {\textbf{0.82}} &{ \textbf{0.87}} & \textbf{0.7} & \textbf{0.74} & {\textbf{0.87}}\\
\hline
{TadGan\cite{geiger2020tadgan} } & 0.76 & 0.76 & {0.69} & 0.58 & 0.58 & {0.68}\\
\hline
{Arima\cite{pena2013anomaly}} & 0.52 & 0.61 & 0.57 & 0.49 & {0.63} & {0.45} \\
\hline
{LSTM AE\cite{hundman2018detecting}} & 0.73 & 0.76 & {0.75} & 0.55 & 0.65 & 0.66\\
\hline
{LSTM \cite{tinawi2019machine}} &0.67  & 0.56 & 0.73 &0.53  & {0.52} & {0.67}\\
\hline
{MCD-BiLSTM-VAE\cite{chen2021imbalanced}} & {0.65} & {0.68} & 0.78 & {0.60} &{ 0.58} &{0.84} \\
\hline
\end{tabular}
}
\end{table}

In Table \ref{T3}, we compare the ${F_1}$ score for the aforementioned  methods for two different baselines.  As the  comparison results show, our proposed method outperforms the existing methods for AD. The closest methods to our Bayesian LSTM method in the $F_1$ score are TadGan, LSTM AE and MCD-BiLSTM-VAE.  The difference between the $F_1$ score of our  Bayesian LSTM method (and also other methods), and the MadGan approach is significant. For this reason,  we have excluded the  MadGan approach from the rest of the  comparisons. We notice that applying the proposed  post-processing approach in Section IV.C to MCD-BiLSTM-VAE method significantly improves the $F_1$ score of the original method in \cite{chen2021imbalanced}; thus we use a modified version of  \cite{chen2021imbalanced} in our comparisons. In addition to $F_1$ score improvement, our method still has other advantages over TadGan, LSTM AE, LSTM, and MCD-BiLSTM-VAE approaches, which have been discussed in Section I. One of the distinct advantages is the complexity reduction of our Bayesian approach compared to other methods  (e.g., TadGan).

  Now let us examine the complexity cost of different methods in detail. Table \ref{T4} shows the computational cost of  our proposed approach compared with other methods. The reported results are performed using the  free Google Colaboratory platform with 2-Core Xeon 2.2GHz, 13GB RAM 
   and the PyTorch framework installed. The comparison results show that the Arima method performs best regarding computational complexity. While the computational complexity of our method is significantly lower than that of the TadGan approach, it is almost similar to the Arima method. Furthermore, the training time of our method is identical to the LSTM approach.  This is because the MC dropout does not affect the NN during the training phase. Instead, during the testing phase, applying the dropout enables us to  approximate the original BNN. Moreover,  the complexity cost of MCD-BiLSTM-VA method is higher than our approach. 
    
Furthermore, in Table \ref{tableV},  we compare the precision, recall,  and accuracy of our method with other approaches. Considering the trade-off between complexity and performance measures, the results of the comparison show, our method outperforms the various competitors.

\section{Conclusions}
The proposed NN-based method provides a generic low-complexity, and prediction-based solution to advancing AD. Our method improves the MSE, precision, recall, accuracy, and $F_1$ score compared to existing methods. The techniques proposed for LSTM can be considered a generic time series analytics tool for autonomous FDIR tasks of satellite missions. Equally important, the proposed method addresses a critical challenge in FDIR: how to efficiently use the time series data without labels to facilitate FDIR missions. The proposed work enables efficient utilization and analysis of the time series data for satellite telemetry and other FDIR applications where on-board hardware and software AD is needed.

\bibliographystyle{IEEEtran}
\bibliography{dropout.bib}

\begin{IEEEbiographynophoto}{Mohammad Amin Maleki Sadr}
He received his  B.Sc. degree from
University of Isfahan, Iran, and the M.Sc., and Ph.D. degrees
 from  K.N. Toosi University of Technology, Tehran, Iran in  2012, 2014, and 2018, respectively,  all in Electrical Engineering. He is now a postdoc research fellow at  Department of Statistics \& Actuarial Science in University of Waterloo, Canada. 
 His research interests include deep learning, anomaly detection, localization, signal processing, and wireless communication.
\end{IEEEbiographynophoto}
\vskip 0pt plus -1fil

\begin{IEEEbiographynophoto}{Yeying Zhu} received her Ph.D. degree in Statistics from Pennsylvania State University, USA. She is currently an Associate Professor at Department of Statistics \& Actuarial Science from University of Waterloo. Her research interest lies in the interface between causal inference and machine learning methods. 
\end{IEEEbiographynophoto}
\vskip 0pt plus -1fil

\begin{IEEEbiographynophoto}{Peng Hu}
received his Ph.D. degree in Electrical Engineering from Queen's University, Canada. He is currently a Research Officer at the National Research Council Canada and an Adjunct Professor at the University of Waterloo. He has served as an associate editor of the Canadian Journal of Electrical and Computer Engineering, a member of the IEEE Sensors Standards committee, and on the organizing and technical committees of industry consortia and international conferences/workshops at IEEE ICC'23, IEEE PIMRC'17, IEEE AINA'15, etc. His current research interests include satellite-terrestrial integrated networks, autonomous networking, and industrial Internet of Things systems.
\end{IEEEbiographynophoto}

\end{document}